\pdfoutput=1
\documentclass[11pt]{article}

\usepackage{acl}

\usepackage{times}
\usepackage{latexsym}
\usepackage{amssymb}
\usepackage{booktabs}
\usepackage{amsfonts}
\usepackage{caption}
\usepackage{threeparttable}
\usepackage{geometry}
\usepackage{multirow}
\usepackage{microtype}
\usepackage{booktabs}
\usepackage{algorithm}
\usepackage[noend]{algpseudocode}
\usepackage{subcaption}
\usepackage{amsmath}
\usepackage{graphicx}
\usepackage{multicol}
\usepackage{xspace}
\usepackage{bm}
\usepackage[T1]{fontenc}
\usepackage{pifont}% http://ctan.org/pkg/pifont
\newcommand{\xmark}{\ding{55}}%
\newcommand*{\affmark}[1][*]{\textsuperscript{#1}}
\makeatletter
\def\thanks#1{\protected@xdef\@thanks{\@thanks
        \protect\footnotetext{#1}}}
\makeatother

\usepackage[T1]{fontenc}
\usepackage[utf8]{inputenc}

\usepackage{microtype}
\definecolor{gred}{RGB}{255,102,102}
\definecolor{gblue}{RGB}{51,102,255}
\definecolor{gyellow}{RGB}{244,180,0}
\definecolor{ggreen}{RGB}{15,157,88}
\definecolor{ggrey}{RGB}{115,115,115}
\definecolor{na}{gray}{0.9}
\definecolor{LightYellow}{RGB}{255,255,191}
\definecolor{LightGreen}{RGB}{152,255,152}
\definecolor{LightRed}{RGB}{255, 204, 203}

\definecolor{OrangeRed}{rgb}{1.0, 0.27, 0.0}
\definecolor{midnightgreen}{rgb}{0.0, 0.29, 0.33}
\definecolor{darkgreen}{rgb}{0.0, 0.42, 0.24}
\definecolor{skyblue}{RGB}{70, 130, 180}

\usepackage{tablefootnote}

\usepackage{tabularx,booktabs,ragged2e}
\usepackage{inconsolata}

\title{Towards Robust Temporal Reasoning of Large Language Models \\
via a Multi-Hop QA Dataset and Pseudo-Instruction Tuning}

\author{Qingyu Tan\thanks{$^{*}$Qingyu Tan is under the Joint PhD Program between Alibaba and NUS.} \affmark[$^{*}$ 1, 2]~~~\textbf{Hwee Tou Ng\affmark[$^\dag$ 2] \thanks{$^\dag$  Corresponding author.}~~~Lidong Bing\affmark[1] } 
\\$^1$DAMO Academy, Alibaba Group~~\\
$^2$Department of Computer Science, National University of Singapore\\
\texttt{\{qingyu.tan,l.bing\}@alibaba-inc.com}\\
\texttt{\{qtan6,nght\}@comp.nus.edu.sg}\\
}

\begin{document}

\maketitle

\begin{abstract}
Knowledge in the real world is being updated constantly. However, it is costly to frequently update large language models (LLMs). Therefore, it is crucial for LLMs to understand the concept of temporal knowledge. However, prior works on temporal question answering (TQA) did not emphasize multi-answer and multi-hop types of temporal reasoning. In this paper, we propose a complex temporal question-answering dataset \textbf{Complex-TR} that focuses on multi-answer and multi-hop temporal reasoning. Besides, we also propose a novel data augmentation strategy to improve the complex temporal reasoning capability and robustness of LLMs. We conducted experiments on multiple temporal QA datasets. Experimental results show that our method is able to improve LLMs' performance on temporal QA benchmarks by significant margins\footnote{Our code and data are released at \url{https://github.com/nusnlp/complex-tr}}.
\end{abstract}

\section{Introduction}
\label{sec:introduction}
Time is a fundamental aspect of the real world. Much information comes with an expiry date. Recent advances of large language models (LLMs) (\citealp{wei2022finetuned,ouyang2022training,achiam2023gpt}) have demonstrated that LLMs can tackle many NLP tasks in a few-shot manner. However, preliminary studies showed that one of the key drawbacks of existing LLMs is the lack of temporal reasoning capability (\citealp{chen2021dataset}; \citealp{tan2023benchmarking}). The \textbf{SituatedQA} \citep{zhang-choi-2021-situatedqa} dataset was first proposed to incorporate extra-linguistic contexts to QA, which include temporal contexts and geographical contexts. \citet{chen2021dataset} proposed the \textbf{TimeQA} dataset and formulated temporal QA as an open-book QA task. \citet{liska2022streamingqa} proposed the \textbf{StreamingQA} dataset by WMT data from 2007 to 2020. \citet{10.1162/tacl_a_00459} constructed the \textbf{TempLAMA} dataset by the Wikidata Knowledge Base with facts from 2010 to 2020. \citet{tan2023benchmarking} proposed a temporal QA benchmark \textbf{TempReason} with coverage of long durations and divided temporal reasoning into three levels: time-time reasoning (\textbf{L1}), time-event (\textbf{L2}) reasoning, and event-event (\textbf{L3}) reasoning. The temporal question-answering (TQA) task is essentially answering questions with temporal constraints, and the answers to the questions are derived from time-dependent facts. An example query is $(s, r, ?, t_{r})$, where $s$ is the subject, $r$ is the relation, $t_{r}$ is the reference time for this question, and the answer denoted by $?$ is the object.
%\citet{zhou-etal-2019-going} created the \textbf{MC-TACO} dataset to examine commonsense reasoning for events' temporal duration. 

\begin{figure}
    \centering
    \resizebox{0.9\columnwidth}{!}{
    \includegraphics{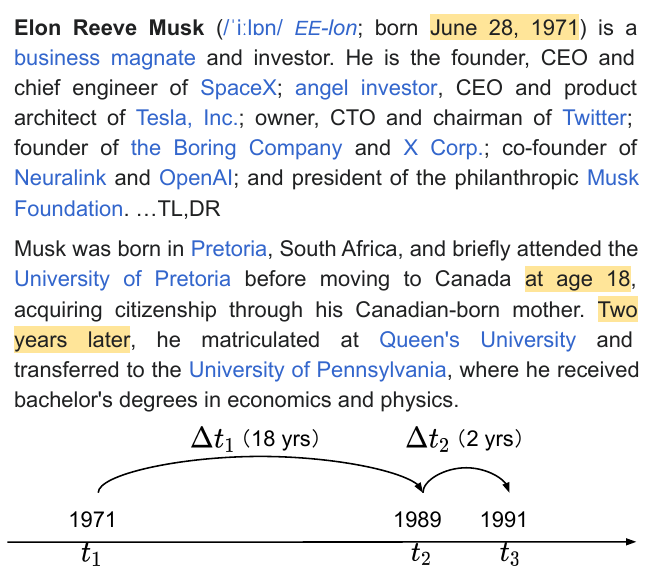}}
    \caption{An example of a 3-hop temporal expression for $t_{3}$. The temporal expressions are highlighted in yellow in the paragraph. The temporal expressions include exact timestamps and time intervals. This example is taken from Elon Musk's Wikipedia page on 18 June 2023.}
    \label{fig:multi-hop-illustration}
\end{figure}

However, existing temporal question-answering datasets have several common drawbacks. The first drawback is that they fail to examine the temporal co-occurrence phenomenon. In the real world, multiple events can happen at the same time, and temporal questions can have multiple valid answers. For example, in Figure~\ref{fig:multi-hop-illustration}, we can see that Elon Musk is the chief executive officer of both Tesla and SpaceX as of June 2023. Nevertheless, all prior benchmarks followed the SQuAD \citep{rajpurkar-etal-2016-squad} evaluation metrics, i.e., token-level F1 and exact match (EM) score. These two metrics take the max scores among all answers when there are multiple answers. Such metrics overestimate the performance of temporal QA and cannot properly evaluate questions with multiple answers. 

The second drawback of existing temporal question-answering benchmarks is that the questions mainly focused on one-hop temporal reasoning, i.e., only one temporal expression is included in the question. For example, in the question ``What team did Kobe Bryant play for in June 2010?'', the temporal constraint refers to only one timestamp. In this paper, we define multi-hop temporal questions as questions that contain multiple temporal expressions. The temporal expression can be a timestamp $t$ or a time interval $\Delta t$. In the real world, temporal concepts are often expressed by multiple time expressions. For example, in Figure \ref{fig:multi-hop-illustration}, $t_{3}$ refers to the year 1991, and it can be explicitly expressed as a numerical value (1-hop expression) or implicitly as $t_1 + \Delta t_{1} + \Delta t_{2}$, as shown in the paragraph. An example question for multi-hop temporal reasoning is ``When did Elon Musk move to Canada?''. In this case, $t_{1}$ is the birth date of Elon Musk and $\Delta t_{1}$ is 18 years. Multi-hop temporal expressions are common in the real world, whereas the study of multi-hop temporal reasoning is under-explored by prior temporal QA datasets. Note that the ``multi-hop'' concept in this paper refers to temporal hops (number of temporal expressions in a question) and they are different from the graphical hops used in QA over knowledge graphs (KGQA) (\citealp{lin-etal-2018-multi}; \citealp{saxena-etal-2020-improving}) and temporal knowledge graphs (\citealp{bai2021multi}; \citealp{bai2023multi}), where the number of graphical hops refers to the number of triples required to answer the question. 
%Even if a question is only relevant to one knowledge triple (one graphical hop), the temporal expressions in the question can be multi-hop. Therefore, we believe it is crucial to distinguish the concept of temporal hops on the time axis from graphical hops in a KG. Since the focus of this paper is studying complex temporal reasoning, the multi-hop questions in this paper refer to multi-hop temporal questions.
To the best of our knowledge, we are the first to differentiate temporal hops from KG hops for temporal question answering.

To address the two shortcomings of the existing datasets, we created a temporal QA dataset \textbf{Complex TempReason} (\textbf{Complex-TR}) that emphasizes multi-hop and multi-answer temporal reasoning. We follow the logical breakdown of the TempReason dataset and focus on time-event (L2) and event-event (L3) reasoning, since these two reasoning types require grounding events to the time axis and are much more challenging than time-time (L1) reasoning. Besides the knowledge from Wikidata KB and Wikipedia articles, our dataset also includes external contexts from Google Custom Search API\footnote{\url{https://developers.google.com/custom-search/v1}} for the open-domain QA (ODQA) setting. To examine the robustness of temporal reasoning, we only used questions before 2020/01/01 as training data. The examples after 2020/01/01 will be used as unseen future questions. Moreover, for our test set, we engaged college-educated human annotators to verify the correctness of the QA pairs. This human verification process ensures that our test set is of high quality for temporal QA research.

%The other major challenge for temporal reasoning is that LLMs tend to be biased towards recent years \citep{tan2023benchmarking}. This is primarily because real-world data are highly imbalanced.
Besides the proposed dataset, we also proposed two methods to improve the performance of temporal QA. The first is Pseudo-Instruction Tuning (PIT), a data augmentation strategy to improve the robustness of temporal reasoning. The second is context refinement, an effective context selection strategy to accommodate long contexts for temporal ODQA. We conducted extensive experiments on our dataset in various TQA settings. Besides, we also conducted experiments on other TQA datasets. Experimental results show that our methods achieve significant performance gains over strong baselines, especially on the out-of-domain years and more complex questions.

In summary, our contributions are as follows:
\begin{itemize}
    \item We are the first to study multi-hop and multi-answer questions for the temporal QA (TQA) task. We also found out that prior benchmarks for TQA adopted inappropriate evaluation metrics for multi-answer questions.
    \item We constructed a complex temporal QA dataset \textbf{Complex-TR} that covers diverse types of multi-hop temporal reasoning by distant supervision and human verification. Experimental results show that all LLMs perform significantly worse on multi-hop temporal questions.
    \item We propose a novel data augmentation strategy to create pseudo-instruction tuning data to improve the complex temporal reasoning capabilities and temporal robustness of LLMs. We also proposed an effective context refinement strategy for the long-context problem in ODQA. Extensive experimental results show that our methods significantly improve the performance over strong baselines.
\end{itemize}
% \usepackage{multirow}

\begin{table*}[ht]
\centering

\resizebox{0.85\textwidth}{!}{
\begin{tabular}{lccccccccc}
\toprule
\textbf{Dataset}           & \textbf{Size}       &\textbf{L2 1-hop} & \textbf{L2 M-hop} & \textbf{L3 1-hop} & \textbf{L3 M-hop} &  
$\Delta \bm{t}$ \textbf{Question} & \textbf{\% M-answer} & \textbf{M-answer eval.} \\ 
\midrule
\textbf{TempLAMA}          & 50K            &    \checkmark      &         \xmark     &    \xmark      &     \xmark         &  \xmark         &         25.3\%        &                  \xmark         \\
\textbf{TimeQA} &  41.2K         &  \checkmark        &       \checkmark       &    \xmark      &  \xmark            &   \xmark         &          6.3\%         &                 \xmark          \\
\textbf{StreamingQA}            & 147K          &   \checkmark       &      \xmark        &     \xmark     &      \xmark        &    \xmark         &  25.0\%             &              \xmark             \\
\textbf{SituatedQA}            & 12.2K         &  \checkmark        &        \xmark      &     \xmark     &  \xmark            &     \xmark         &  4.7\%            &                    \xmark       \\
\textbf{TempReason}              & 52.8K  &  \checkmark        &         \xmark     &     \checkmark     &     \xmark         &       \xmark         &  8.6\%          &   \xmark                      \\
\midrule
\textbf{Complex-TR (Ours)}    &      10.8K         &   \checkmark       &            \checkmark  &   \checkmark       &    \checkmark  &  \checkmark         &    23.5\%     &  \checkmark    \\                        \bottomrule          
\end{tabular}}
\caption{Comparison between Complex-TR and prior temporal QA datasets. The \% M-answer column refers to the percentage of multi-answer questions. We can see that all prior datasets contain a considerable number of multi-answer questions, yet none of them used appropriate evaluation metrics for multi-answer questions.}
\label{tab:dataset-compare}
\end{table*}

\section{Our Dataset}
\label{sec:dataset}

%``What awards did Kobe Bryant win when he was playing for the Los Angels Lakers?''

 \citet{tan2023benchmarking} first proposed to divide temporal reasoning into three levels: time-time reasoning (\textbf{L1}),  time-event reasoning (\textbf{L2}), and event-event reasoning (\textbf{L3}). In this paper, we focus on the harder temporal reasoning types: time-event reasoning (\textbf{L2}) and event-event reasoning (\textbf{L3}). We construct our dataset by the following steps:

 \noindent\textbf{Mining Temporal Facts in Wikidata} We first extract all the knowledge triples that have temporal qualifiers (such as \textit{start\_time} and \textit{end\_time}) in the Wikidata \citep{vrandevcic2014wikidata} knowledge base (2023/03/20 dump). We then reformat the extracted triples into temporal quintuples $(s, r, o, t_{s}, t_{e})$, where $s$ is the subject, $r$ is the relation, $o$ is the object, $t_{s}$ is the start time, and $t_{e}$ is the end time. We then group the quintuples with the same subject together, obtaining $S = \{(s, r_{i}, o_{i}, t_{si}, t_{ei})|i \in 1...N \}$. Unlike prior works from Wikidata (\citealp{chen2021dataset}; \citealp{tan2023benchmarking}), where only one relation type is kept within one group, we include multiple relation types in one group, which adds more complexity to our dataset. For each group, we identify the most common relation as its representative relation. Since the relation distribution is highly imbalanced in the Wikidata KB, we set a ceiling of 250 groups for each representative relation type. In the end, we kept 2,000 temporal quintuple groups, and on average there are 9.2 temporal quintuples in each group. We divide the groups into training (1,000), development (500), and test (500) sets.
 \begin{table*}[ht]
\resizebox{\textwidth}{!}{
\begin{tabular}{lllc}
\toprule
\textbf{Dataset}                            & \textbf{Type} & \textbf{Example Question}                                                                             & \textbf{Answers}                             \\
\midrule
\textbf{TempLAMA}                           & L2 1-hop       & In 2011, Tom Brady played for \_X\_.                                                          & \textbf{New England Patriots}               \\
\textbf{TimeQA}                  & L2 M-hop       & Which team did Olivier Bernard play for from 2000 to 2005?                                   & \textbf{Newcastle United}                   \\
\textbf{StreamingQA}                        & L2 1-hop       & Which player scored for St Mirren in November 2008?                                          & \textbf{Franco Miranda}                     \\
\textbf{SituatedQA}                         & L2 1-hop       & Who made the most free throws in NBA history as of 2020?                                     & \textbf{Karl Malone}                        \\
\multirow{2}{*}{\textbf{TempReason}}        & L2 1-hop       & Where was Barack Obama educated in Apr 1981?                                                 & \textbf{Columbia University}                \\
                                   & L3 1-hop       & Who was the chair of Swedish People's Party of Finland after Lars Erik Taxell?               & \textbf{Jan-Magnus Jansson}                 \\
\midrule
\multirow{4}{*}{\textbf{Complex-TR (Ours)}} & L2 M-hop   & Who were the chairs of FC Barcelona from March 1984 to March 2003?                           & \textbf{Josep Lluís Núñez} and \textbf{Enric Reyna}  \\
                                   & L2 M-hop   & Where was Lynne Bowker educated 15 years before June 2005?                                   & \textbf{University of Ottawa}               \\
                                   & L3 M-hop   & Which employer did Barack Obama work for 2 years after he/she studied at Occidental College? & \textbf{Business International Corporation} \\
                                   & L3 M-hop   & Who were the owners of Chelsea F.C. when Thomas Tuchel was the headcoach?                    & \textbf{Roman Abramovich} and \textbf{Todd Boehly}  \\
\bottomrule
\end{tabular}}
\caption{Example questions of prior temporal QA datasets and our Complex-TR dataset.}
\label{tab:example-compare}
\end{table*}

\noindent\textbf{Creating Questions from Quintuples} After obtaining the temporal quintuple groups, we create the question-answer pairs based on manually designed templates (details are shown in Appendix~\ref{sec:question-templates}). For L2 temporal questions, a 1-hop question can be expressed in the query $(s, r, ?, t_{r})$, where $t_{r}$ is the reference timestamp. We create the multi-hop L2 questions with two variations: (1) $(s, r, ?, t_{rs}, t_{re})$, where $t_{rs}$ and $t_{re}$ refer to the start and end time of the question respectively. (2) $(s, r, ?, t_{r}, \Delta t)$, where $t_{r}$ is the reference time and $\Delta t$ is the temporal difference between the reference time and the query time. The model is expected to infer the query time from $\Delta t$ and $t_{r}$. As for the L3 questions, the number of temporal hops is dependent on the event-event temporal relation. For example, from the question ``What team did Kobe Bryant play for before the LA Lakers?'', we can imply that the query time is the start time of Kobe Bryant playing for the Lakers. Since only one timestamp is implied in the question, it is a 1-hop question. In contrast, for the question, ``What awards did Kobe Bryant win when he was playing for the LA Lakers?'', both the start time and end time have to be considered. Since Kobe Bryant had a 20-year career with the Lakers, it is a multi-hop question. 

In order to have a fair comparison between 1-hop and multi-hop temporal reasoning, we also created a small number of 1-hop questions with the same temporal quintuples and contexts. We denote the timestamp mentioned in the question as the reference time. For L3 questions, we denote the start time of the reference event as the reference time. In order to examine the robustness of temporal reasoning in future years, we only use the questions with a reference time before 2020/01/01 for training, whereas the development and test sets contain questions both before and after 2020/01/01. 

\noindent\textbf{Open Domain Context Retrieval} Previous temporal QA datasets typically rely on a fixed knowledge source. The TimeQA and TempReason datasets use Wikipedia articles as context for open-book QA setting. The StreamingQA dataset uses English news articles from the WMT challenge. TimeQA is a human-annotated dataset based on Wikipedia articles, and temporal facts not reflected in Wikipedia are deemed as ``unanswerable''. However, a single knowledge source may not be sufficient to answer temporal questions, as temporal facts are constantly evolving. To address this limitation, we construct our dataset in an open-domain QA (ODQA) fashion. For the context used in the ODQA setting, we first include the Wikipedia article on the question subject. 
We then use the Google Custom Search API with this question as the search query to retrieve the top 10 results. The searched web pages will be scraped by Trafilatura~\cite{barbaresi-2021-trafilatura}, a state-of-the-art text extraction tool for NLP research. Due to the anti-scraping mechanism of certain websites, we cannot extract the contexts from all the retrieved web pages. On average, we have 6.7 articles for each question and the average total context length is 93K words. The long context from multiple sources introduces additional challenges for the temporal QA task.

\noindent\textbf{Human Verification} To ensure that our dataset is of high quality and aligned with the retrieved contexts, we engaged college-educated human annotators to verify the correctness of the QA pairs and retrieved contexts. The annotators are given a temporal QA pair and its corresponding articles, and they are asked to read through the articles and judge whether the QA pair is correct or not. Due to cost limitations, we randomly sample 500 QA pairs from the test set for human verification. Among the 500 QA pairs, 100 of them are annotated by two annotators to measure the inter-annotator agreement. The Fleiss Kappa coefficient of this subset is 0.71, which implies a substantial agreement level. The conflicts are resolved by a third annotator.  The hourly pay of the annotators is above 22 USD, which is significantly higher than the local minimum wage. In the end, 329 QA pairs are deemed as correctly reflected in the contexts, we then used these 329 QA pairs as our gold test set in the experiments. We name our dataset Complex-TempReason (\textbf{Complex-TR}). Our detailed dataset statistics are shown in Table~\ref{tab:data-stats}. The``Pseudo'' column refers to the pseudo-data that we generated in Section~\ref{sec:pit}. Besides, we also compare our questions and reasoning types with prior temporal QA datasets in Table~\ref{tab:dataset-compare}. We can see that none of the prior datasets included L3 multi-hop temporal reasoning, or questions with time interval $\Delta t$. Besides, all prior temporal QA datasets contain a considerable number of questions with multiple answers, yet none of them adopted any evaluation metrics for multi-answer questions. We provide a more detailed question comparison of our dataset and prior works in Table~\ref{tab:example-compare}.
\begin{table}[ht]
\resizebox{\columnwidth}{!}{
\begin{tabular}{lcccc}
\toprule
                    & \textbf{Pseudo}    & \textbf{Training}     & \textbf{Dev}       & \textbf{Test (Gold)}      \\
\midrule
\textbf{Time Coverage}       & 1021-2040 & 1529-2019 & 1254-2023 & 1659-2023 \\
\textbf{L2 1-hop}            &11,389
                 & 1,109      & 670       & 72       \\
\textbf{L2 M-hop}            & 13,919
       & 2,407      & 1,487      & 106     \\
\textbf{L3 1-hop}            & 15,205 & 1,236      & 690       & 71       \\
\textbf{L3 M-hop}            & 10,938 & 1,759      & 1,146      & 80       \\
\textbf{\#Total}          & 51,451  & 6,511     & 3,993       & 329       \\
%\textbf{\#Subjects}          & 3,363  & 1,000     & 500       & 500      \\
%\textbf{facts/subject} & 10.8 & 9.0         & 10.0        & 8.6     \\
%\textbf{facts/subject} & 10.8 & 9.0         & 10.0        & 8.6     \\
\textbf{Avg. Facts} & 10.8 & 9.0         & 10.0        & 8.6     \\
\textbf{Avg. Contexts} & - & 6.6         & 6.8        & 6.7     \\
\textbf{Avg. Word Len.} & - & 92K         & 99K        & 94K     \\

\bottomrule
\end{tabular}}
\caption{Statistics of our dataset. Note that we do not include questions after December 2019 in our training set. The \textbf{Avg. Word Len.} row refers to the average number of words in all contexts for the ODQA setting.}
\label{tab:data-stats}
\end{table}
\section{Methodology}
In this paper, we also propose two strategies to improve the robustness and the capabilities of LLMs for temporal reasoning, which are Pseudo-Instruction Tuning and Context Refinement. Pseudo-Instruction Tuning uses pseudo data to alleviate the data scarcity and data imbalance problems of temporal QA, which can improve LLMs' temporal reasoning capability and robustness. The improved reasoning capability also has a positive impact in the ODQA setting. On the other hand, Context Refinement is used to address the long context problem in the ODQA setting. In this setting, the total context can reach over 100K tokens, which makes it infeasible to feed the context to existing QA models. The Fusion-in-Decoder (FiD;~\citealp{izacard-grave-2021-leveraging}) model was proposed for the QA task to process long contexts. It breaks a long context into multiple smaller paragraphs to avoid the quadratic computation of self-attention. However, even FiD can only process up to 10K tokens, which is still insufficient for ODQA. As such, we use sentence embedding models to refine the paragraphs and only keep the most relevant paragraphs to the question for our QA models. We will introduce each strategy in detail in the following subsections.
\subsection{Pseudo-Instruction Tuning}
\label{sec:pit}
One of the main challenges of temporal QA is that the labeled data are typically concentrated on recent years, and some datasets (\citealp{liska2022streamingqa}; \citealp{10.1162/tacl_a_00459}) only contain data from 2000-2020. The data imbalance is caused by data distribution in the Wikidata knowledge base. As a result, LLMs trained by such data tend to be biased towards recent years \citep{tan2023benchmarking}. To overcome this challenge, we aim to create artificial data with an emphasis on low-frequency time periods.

\noindent\textbf{Pseudo-Data Generation} Data augmentation has proven to be effective in many NLP tasks (\citealp{zhou2021melm}; \citealp{ding-etal-2020-daga}; \citealp{cao-etal-2023-mitigating}). For TQA, we have the training group $S$ of subject $s$ as:
%\vspace{-0.8mm}
\begin{equation}
    S=\{(s, r_{i}, o_{i}, t_{si}, t_{ei})|i \in 1...N \}
%\vspace{-0.8mm}
\end{equation}
  We then shift $S$ by $\Delta t$ for every fact within that group, obtaining:
  \begin{equation}
  \resizebox{0.85\columnwidth}{!}{
      $S_{p} = \{(s, r_{i}, o_{i}, t_{si} + \Delta t, t_{ei} +\Delta t)|i \in 1...N \}$
      }
  \end{equation}
 where $-100  \leq \Delta t \leq +20$ is a random temporal shift with a maximum of 20 years going forward and a maximum of 100 years going backward, and $S_{p}$ is the shifted pseudo-group. Since shifting temporal facts introduces temporally augmented facts, we replace all the subjects and objects with fictional entities to avoid conflicts. We used ChatGPT to generate multiple types of fictional entities, such as person names and sports teams. This process can be repeated multiple times. In this way, we can create large amounts of artificial temporal quintuples. We then follow the question templates in Section \ref{sec:dataset} to generate question-answer pairs using fictional facts, and hence generate temporal reasoning data without human annotation. Examples of fictional entities and fictional temporal facts are shown in Appendix~\ref{sec:fiction-explanations}.

%The large-scale pseudo data are used to pre-train LLMs, the questions and the fictional contexts will be prompted to LLMs, and the model is expected to generate the corresponding answers. 

\noindent\textbf{Temporal Resampling} To accommodate data imbalance in the Wikidata KB, especially for future years, we resample the generated pseudo-data by time intervals. Specifically, we divide time into 20-year intervals from 1900 to 2020, and then count all the examples in our training dataset within each time interval, obtaining $\{ n_{i}| i \in 1..k\}$. Examples before 1900 will be treated as one group, so there are 7 counts in total and $k=7$. Note that our training data does not contain any questions after December 31, 2019. Hence, examples after December 31, 2019 in the development and test sets will be used to simulate future data. We calculate probabilities for resampling by:
\begin{equation}
    P_{i} = 1 - \frac{n_{i}}{max(\{ n_{i}| i \in 1..k\})}
\end{equation}
We then sample the generated questions with probability $P_{i}$. For pseudo-data after 2019, we set the sampling probability to be 1. In this way, we will be able to de-bias the temporal distribution in the Wikidata KB and let the models focus on improving the performance on low-frequency years. 

\noindent\textbf{Training for PIT} After we obtained the resampled pseudo-data, we followed the instruction templates for QA tasks from FLAN \citep{wei2022finetuned} to fine-tune the LLMs. We name this process Pseudo-Instruction Tuning (\textbf{PIT}). The final size and statistics of PIT are shown in the ``Pseudo'' column in Table~\ref{tab:data-stats}. The instruction-tuned LLMs will then be used to fine-tune on the task-specific data. We believe that improving temporal reasoning capability can have a positive impact on downstream tasks such as ODQA.

\subsection{Context Refinement}
\label{sec:reranker}
In the ODQA setting, the contexts are from multiple sources. We first denote our context set as $C$, and follow the pre-processing protocol of FiD to split all the articles into 100-word paragraphs:
\begin{equation}
    C = \{p_{1}, p_{2}, p_{3}, ..., p_{m} \}
\end{equation}
where $p_{i}$ refers to a paragraph in the context set. We then encode the temporal question $q$ and the paragraphs by a sentence embedding model $f$ and then calculate their cosine similarity:
\begin{equation}
    z_{i} = Cos(f(q), f(p_{i}))
\end{equation}
$z_{i}$ is used as a relevance score to re-rank all the paragraphs. Due to computation constraints, we only use the top $k$ paragraphs as the contexts for the FiD model. In this way, we can refine the extra-long context to an acceptable level. We examined multiple sentence embedding models and chose \texttt{bge-base-en-v1.5} \cite{bge_embedding} as our sentence encoder $f$. It is an advanced embedding model on the MTEB~\cite{muennighoff-etal-2023-mteb} benchmark and works best in our experiments We show the ablation studies of different re-rankers in Appendix~\ref{sec:ablation}.
\section{Experiments}

\subsection{Experimental Setup}

In this paper, we focus on two temporal QA settings. The first is open-domain QA (\textbf{ODQA}), where the models are provided with multiple retrieved articles as context. The second is the \textbf{ReasonQA} setting proposed by \citet{tan2023benchmarking}, where all the relevant structured knowledge quintuples to answer a question are provided as context. This is because we aim to study the reasoning aspect of temporal QA. We elaborate on the problem settings in Appendix \ref{sec:prob-setting-example} in greater detail.

\noindent\textbf{Baselines} (1) \textbf{FLAN-T5-XL} \citep{wei2022finetuned} This model is an instruction-tuned encoder-decoder model with 3B parameters. It achieves respectable few-shot performance on the MMLU benchmark. (2) \textbf{GPT-3.5} \citep{ouyang2022training} We used \texttt{gpt3.5-turbo} as our baseline. (3) \textbf{GPT-4} \citep{achiam2023gpt} This model is the most advanced LLM in the market. It achieves strong zero-shot performance on many NLP tasks. Since the model is constantly being updated, we used the \texttt{gpt4-0613} model for consistent evaluation. We evaluate the one-shot performance of the first three LLMs. (4) \textbf{T5-SFT} \citep{raffel2020exploring} This model is the supervised fine-tuned model with our labeled training data. We used the T5 models as our backbone model. We conducted experiments on T5-base (\textbf{T5-B}) and T5-large (\textbf{T5-L}). In the ODQA setting, we truncate the context to 1,024 tokens due to GPU memory constraint. (5) \textbf{T5-PIT-SFT} This model is first instruction-tuned by pseudo data and then further fine-tuned with labeled data. (6) \textbf{FiD} is the supervised fine-tuned baseline for FiD. It uses T5 as its backbone model and splits a long context into smaller paragraphs. Hence, the maximum context window of FiD is significantly larger than T5 under the same memory constraint. (7) \textbf{FiD-PIT} combines T5-PIT initialization with FiD training strategy. (8) \textbf{FiD-PIT-Refined} is the FiD-PIT model with context refinement described in Section~\ref{sec:reranker}. For SFT models, we report the average result of three random runs.

\subsection{Evaluation Metrics}

As mentioned in Section \ref{sec:introduction}, all of the prior TQA benchmarks followed SQuAD \citep{rajpurkar-etal-2016-squad}. However, the metrics in SQuAD are computed by using the maximum scores with all references, which significantly overestimate the performance for multi-answer questions. Therefore, we adopted two stricter metrics for our experiments. The first metric is set-level accuracy (\textbf{Set Acc.}; \citealp{zhong2022romqa}). This metric will only return correct if the prediction set is identical to the ground truth set. The second additional metric is answer-level F1 (\textbf{Ans. F1}; \citealp{amouyal2022qampari}). Unlike token-level F1 in SQuAD, \textbf{Ans. F1} counts true positives only when there is an exact match in the answer set. Besides, if the prediction contains extra answers, it will also be penalized. The upper bound of these two metrics is the EM score. We also further analyze the four metrics in Section~\ref{sec:compare-metrics}.

\begin{table}[ht]
\resizebox{\columnwidth}{!}{
\begin{tabular}{lcccc}
\toprule
           & \multicolumn{2}{c}{\textbf{Single-hop}} & \multicolumn{2}{c}{\textbf{Multi-hop}} \\
           & \textbf{Set Acc.}        & \textbf{Ans. F1}       & \textbf{Set Acc.}       & \textbf{Ans. F1}       \\
\midrule
\textbf{FLAN-T5-XL} & 61.5           & 64.1          & 35.5          & 49.7          \\
\textbf{GPT-3.5}    & 28.0             & 45.3          & 31.2          & 51.8          \\
\textbf{GPT-4}      & 67.1           & 80.2          & 51.6          & 65.4          \\
\midrule
  \textbf{T5-\textit{base}}         &                &               &               &               \\
\midrule
\textbf{SFT}        & 80.4           & 83.3          & 59.1          & 65.1          \\
\textbf{PIT-SFT (Ours)}    & \textbf{91.6}           & \textbf{93.8}          & \textbf{78.0}            & \textbf{82.4}          \\
\midrule
  \textbf{T5-\textit{large}}          &                &               &               &               \\
\midrule
\textbf{SFT}        & 86.0             & 88.1          & 71.0            & 76.4          \\
\textbf{PIT-SFT (Ours)}    & \textbf{95.1}           & \textbf{95.6}          & \textbf{85.0}            & \textbf{89.5} 
\\ 
\bottomrule
\end{tabular}}

\caption{ReasonQA experimental results (in \%) that compare single-hop and multi-hop temporal reasoning. The context used in this setting is structured facts in KB.}
\label{tab:mhop-exp}
\end{table}

\begin{table}[ht]
\resizebox{\columnwidth}{!}{
\begin{tabular}{lcccc}
\toprule
              & \multicolumn{2}{c}{\textbf{Single-Hop}} & \multicolumn{2}{c}{\textbf{Multi-hop}} \\
              & \textbf{Set Acc.}        & \textbf{Ans. F1}       & \textbf{Set Acc.}       & \textbf{Ans. F1}       \\
\midrule
\textbf{FLAN-T5-XL}    & 30.1           & 31.5          & 14.5          & 20.3          \\
\textbf{GPT-3.5}        & 17.5           & 26.3          & 9.7           & 23.0            \\
\textbf{GPT-4}          & 19.6           & 35.1          & 14.0            & 37.2          \\
\midrule
    \textbf{T5-\textit{base}}          &                &               &               &               \\
\midrule
\textbf{SFT}           & 33.6           & 35.6          & 17.7          & 26.6          \\
\textbf{PIT-SFT (Ours)}       & 39.9           & 40.8          & 22.6          & 30.1          \\
\textbf{FiD}         & 33.6           & 35.6          & 17.7          & 26.0            \\
\textbf{FiD-PIT (Ours)}     & 39.2           & 40.1          & 23.1          & 30.2          \\
\textbf{+Refine (Ours)} & \textbf{42.7}           & \textbf{44.3}          & \textbf{24.7}          & \textbf{31.2}          \\
\midrule
  \textbf{T5-\textit{large}}            &                &               &               &               \\
\midrule
\textbf{SFT}           & 35.0             & 35.7          & 23.1          & 32.1          \\
\textbf{PIT-SFT (Ours)}       & 39.2           & 40.1          & 25.3          & 32.8          \\
\textbf{FiD}         & 46.2           & 46.6          & 27.4          & 37.3          \\
\textbf{FiD-PIT (Ours)}     & 46.9           & 47.8          & 29.0            & 37.5          \\
\textbf{+Refine (Ours)} & \textbf{49.0}             & \textbf{49.7}          & \textbf{31.2}          & \textbf{39.1}  
\\
\bottomrule

\end{tabular}}
\caption{ODQA experimental results (in \%). The contexts used in this setting are multiple web articles.}
\label{tab:odqa_exp}
\end{table}

\subsection{Experimental Results}

In Table~\ref{tab:mhop-exp}, we can see that multi-hop temporal reasoning is much more challenging compared to single-hop reasoning, and most tested models have lower performance on multi-hop temporal reasoning. This shows that multi-hop temporal reasoning is a common weakness for current LLMs. The GPT-4 model can achieve relatively good results for the answer F1 metric, whereas its set accuracy scores are still significantly below our supervised models. This indicates that GPT-4 can find valid answers under time constraints, but it cannot find all correct answers when multiple answers are presented. We provided the example errors by GPT-4 in Appendix~\ref{sec:error-analysis}.
%Moreover, our final model \textbf{T5-B-PIT-SFT} performs significantly better than \textbf{T5-B-SFT}, especially on multi-hop temporal reasoning. For L2 multi-hop reasoning, our model is able to outperform \textbf{T5-B-SFT} by 10.6 Answer F1 (80.0 vs 69.4). In the L3 M-hop scenario, our model outperforms \textbf{T5-B-SFT} by 13.3 Answer F1 (76.7 vs 63.4).

The experimental results of open-domain QA are shown in Table~\ref{tab:odqa_exp}. The ODQA setting is a much harder QA setting. We can see that all models perform significantly worse than in the ReasonQA setting. Nevertheless, our \textbf{PIT-SFT} and \textbf{FiD-PIT} models still outperform their corresponding baselines (\textbf{SFT} and \textbf{FiD}) significantly. This experimental result verified our assumption that improving temporal reasoning capability can also improve the downstream performance of ODQA. We also show the experimental results of \textbf{PIT} on the TimeQA dataset in Appendix~\ref{sec:further-exp}. We find that \textbf{PIT} also has a positive impact on TimeQA, which demonstrates the generalizability of \textbf{PIT}. 

Since our dataset contains multiple articles for ODQA, the total context length can easily exceed FiD's limit. However, with our proposed context refinement strategy, we can further improve our best model \textbf{FiD-PIT} (large) by 2.1\% in set accuracy for single-hop questions and 2.2\% in set accuracy for multi-hop questions

\begin{table}[ht]
\resizebox{\columnwidth}{!}{
\begin{tabular}{lccc}
\toprule
           & \textbf{In-domain} & \textbf{Future} & \textbf{Overall} \\
           &  & \textbf{Set Acc.} & \\
\midrule
\textbf{FLAN-T5-XL} & 46.5      & 54.5   & 46.8    \\
\textbf{GPT-3.5}    & 30.2      & 18.2   & 28.0    \\
\textbf{GPT-4}      & 58.2      & 63.6   & 58.4    \\
\textbf{T5-B}       & 69.2      & 45.5   & 68.4    \\
\textbf{T5-B-PIT}   & 84.9      & 54.5   & 83.9    \\
\textbf{T5-L}       & 78.0      & 63.6   & 77.5    \\
\textbf{T5-L-PIT}   & \textbf{89.3}      & \textbf{90.9}   & \textbf{89.4}   \\
\bottomrule

\end{tabular}}
\caption{Analysis of the ReasonQA performance for the in-domain years (before 2020/01/01) and out-of-domain future years. The numbers reported in this table are the \textbf{Set Acc.} scores.}
\label{tab:year-analysis}
\end{table}
\section{Analysis}
\label{sec:analysis}
\subsection{Robustness of Temporal Reasoning}
\label{sec:temporal-robustness}

In this section, we analyze temporal reasoning performance by time periods. In an ideal scenario, temporal reasoning capability should generalize to unseen time periods. In Table~\ref{tab:year-analysis}, we show the experimental results for the in-domain and the out-of-domain (OOD) subsets. We treat the questions after 2020/01/01 as OOD (future examples). We believe that this is a valid assumption because our training data do not contain such questions and our backbone model T5 was released in October 2019. From Table~\ref{tab:year-analysis}, we can see that \textbf{FLAN-T5-XL} and \textbf{GPT-4} have higher performance on the future subset. This could be because these two models are fine-tuned on data after 2020. For the supervised models, we can see that \textbf{T5-B}, \textbf{T5-B-PIT}, and \textbf{T5-L} all suffered from severe performance degradation on the future subset, whereas \textbf{T5-L-PIT} can achieve similar performance on in-domain years and future years. This implies that achieving robust temporal reasoning not only requires de-biased pseudo-data but also a good capability of the base language model.

\begin{table}[ht]
\resizebox{\columnwidth}{!}{
\begin{tabular}{lcccc}
\toprule
           & \multicolumn{4}{c}{\textbf{Single-Answer}}    \\
           & \textbf{Set Acc.} & \textbf{Ans. F1} & \textbf{EM}   & \textbf{Tok. F1} \\
\midrule
\textbf{FLAN-T5-XL} & 59.5    & 59.5    & 59.9 & 70.7   \\
\textbf{GPT3.5}     & 29.0    & 48.5    & 53.7 & 62.7   \\
\textbf{GPT-4}      & 60.6    & 71.2    & 73.8 & 79.0   \\
\textbf{T5-B}       & 71.8    & 72.1    & 72.2 & 78.0   \\
\textbf{T5-B-PIT}   & 86.5    & 87.0    & 86.5 & 89.0   \\
\textbf{T5-L}       & 79.9    & 81.2    & 80.7 & 84.4   \\
\textbf{T5-L-PIT}   & \textbf{90.7}    & \textbf{91.8}    & \textbf{91.5} & \textbf{93.9}   \\
\midrule
           & \multicolumn{4}{c}{\textbf{Multi-Answer}}    \\
           & \textbf{Set Acc.} & \textbf{Ans. F1} & \textbf{EM}   & \textbf{Tok. F1} \\
\midrule
\textbf{FLAN-T5-XL} & 0.0     & 43.1    & 65.7 & 81.7   \\
\textbf{GPT3.5}     & 32.9    & 50.8    & 62.9 & 75.2   \\
\textbf{GPT-4}      & 50.0    & 74.2    & 91.4 & 92.7   \\
\textbf{T5-B}       & 55.7    & 76.5    & 87.1 & 91.5   \\
\textbf{T5-B-PIT}   & 74.3    & 88.6    & 94.3 & 96.2   \\
\textbf{T5-L}       & 68.6    & 82.4    & 90.0 & 94.1   \\
\textbf{T5-L-PIT}   & \textbf{84.3}    & \textbf{93.5}    & \textbf{98.6} & \textbf{99.8}  \\

\bottomrule
\end{tabular}}
\caption{Experimental comparison between the performances on single-answer questions and multi-answer questions of selected models. The experiments are conducted on the Complex-TR dataset in the ReasonQA setting.}
\label{tab:compare-metrics}
\end{table}
\subsection{Analysis of Multi-Answer Questions}
\label{sec:compare-metrics}

The ability to understand co-occurring events is a crucial aspect of temporal reasoning. In this section, we analyze the reasoning performance on single-answer and multi-answer questions. We report their experimental results separately in Table~\ref{tab:compare-metrics}. Besides the set-level accuracy (\textbf{Set Acc.}) and answer-level F1 (\textbf{Ans. F1}) reported in the main experiments, we also include exact match (\textbf{EM}) and token-level F1 (\textbf{Tok. F1}). The \textbf{EM} and \textbf{Tok. F1} scores are adopted by all prior temporal QA benchmarks. Both metrics take the maximum scores with all possible answers. In the multi-answer scenario, \textbf{EM} and \textbf{Tok. F1} are generally much higher than \textbf{Set Acc.} and \textbf{Ans. F1}. \textbf{Tok. F1} scores for multi-answer questions are even higher than those of single-answer questions. This indicates that \textbf{EM} and \textbf{Tok. F1} can significantly overestimate the performance of multi-answer questions. Therefore, it is better for temporal QA benchmarks to adopt \textbf{Set Acc.} and \textbf{Ans. F1} for evaluation.

\section{Related Work}

\noindent\textbf{Temporal Information Extraction} Early studies of temporal research in NLP focused on studying temporal relations of short-term events. The TimeBank \citep{pustejovsky2003timebank} dataset was first proposed as a benchmark for the temporal information extraction (TIE) task. It is a human-annotated dataset with annotated events, temporal expressions, and temporal relations (such as \textit{before}, \textit{after}, and \textit{contains}). The TempEval challenges (\citealp{verhagen-etal-2007-semeval}; \citealp{verhagen2010semeval}; \citealp{uzzaman2013semeval}) were later proposed for the TIE task. The schema of TempEval is similar to that of TimeBank. \citet{cassidy-etal-2014-annotation} found that prior TIE datasets were not exhaustively annotated and introduced a dense annotation schema for event ordering. They also released the Timebank-Dense dataset, which is a more complete version of TimeBank. \citet{han-etal-2019-joint} proposed an end-to-end framework to jointly extract events and temporal relations. However, TIE research focused on studying the order of short-term events within a specific context. On the other hand, the focus of our paper is studying temporal reasoning with factual grounding on the global time axis.

\noindent\textbf{Temporal Question Answering} Since time is a fundamental aspect in real-life applications, numerous efforts have been made to study the temporal reasoning problem in question answering (QA). The first line of work in this field worked on QA over temporal knowledge graphs (TKGs). The TempQuestions \citep{Zhen2018TempQuestionsAB} dataset was introduced to extend the KGQA task to TKGs. \citet{Jia2021ComplexTQ} introduced the TimeQuestions dataset as an extension of TempQuestions. The CronQuestions \citep{saxena-etal-2021-question} dataset is a large-scale QA dataset over TKG with more complex questions. \citet{shang-etal-2022-improving} proposed a temporal contrastive learning approach to improve the time-sensitivity for the TKGQA task. However, the TKGQA task is not the focus of our paper, and it requires models to rank all the nodes (including entities and timestamps) for each question. That is, the TKGQA task assumes that all nodes are known to the model, whereas our focus is on performing temporal reasoning over raw natural language text.

The temporal QA task for LLMs was derived from conventional QA. Even though there is a subset of time-related questions in popular QA datasets such as SQuAD \citep{rajpurkar-etal-2016-squad}, the questions are usually asking for a temporal expression present in the context without temporal reasoning. The MC-TACO \citep{zhou-etal-2019-going} dataset was later proposed to study temporal commonsense reasoning. However, MC-TACO did not study the evolving aspect of temporal reasoning, which is crucial for LLMs' continual learning. SituatedQA \citep{zhang-choi-2021-situatedqa} first introduced extra-linguistic context to conventional QA, and it contains evolving temporal questions. \citet{chen2021dataset} proposed the TimeQA dataset with Wikipedia articles and the Wikidata knowledge base. The TempLAMA \citep{10.1162/tacl_a_00459} dataset was later constructed similarly, but it focused on the close-book QA setting. \citet{kasai2023realtime} proposed a real-time QA benchmark that updates questions weekly.  \citet{tan2023benchmarking} systematically tackled the TQA problem by breaking down temporal reasoning into three levels. However, even though most of the previously proposed datasets contain questions with multiple answers, none of them studied multi-answer and multi-hop temporal reasoning. The concept of ``multi-hop'' is commonly used in QA to describe complex questions. The term ``multi-hop'' has different meanings in different contexts. \citet{yang-etal-2018-hotpotqa} used the term to describe questions that can only be answered by multiple paragraphs. In the prior works of KGQA (\citealp{lin-etal-2018-multi}; \citealp{saxena-etal-2020-improving}), ``multi-hop'' describes questions that can only be answered by multiple knowledge triples. Since our paper focuses on temporal reasoning, we use ``multi-hop'' to describe questions that contain multiple temporal expressions. A temporal expression can be a specific timestamp or a time interval.

\section{Conclusions}

In this paper, we studied the under-explored multi-hop temporal reasoning problem in temporal QA. We proposed a novel dataset Complex-TR that covers multi-hop temporal reasoning. Besides, we found out that all prior temporal reasoning benchmarks used inappropriate evaluation metrics (exact match and token F1) for this task. In addition, we proposed Pseudo-Instruction Tuning to enhance the robustness of temporal reasoning and Context Refinement to alleviate the long-context problem in ODQA. Extensive experimental results showed that our methods are significantly better than strong baseline methods.

\section{Limitations}

Since our dataset is constructed from the Wikidata knowledge base, it may retain some errors present in the Wikidata KB. That is, the training and validation sets of our dataset may contain factual errors. However, we ensured the high quality of our gold test set by a rigorous human verification process. The other limitation, for the experiments on Open-domain QA, we leveraged the Wikipedia page and the retrieved results from Google Custom Search API on 2023/12/10. The retrieval results may be different as the internet evolves. Our experimental results are also dependent on the retrieval performance of Google Custom Search API. Nevertheless, it is by far the best-performing retrieval tool for the ODQA task as shown in other QA works (\citealp{kasai2023realtime}; \citealp{zhao-etal-2023-verify}). 

\section{Ethics Statement}

We created our Complex TempReason (Complex-TR) dataset from the Wikidata knowledge base. Wikidata is open-source and under the Creative Commons CC0 License\footnote{\url{https://www.wikidata.org/wiki/Wikidata:Licensing}} and Wikipedia articles are under the Creative Commons AttributionShareAlike 3.0 License\footnote{\url{https://en.wikipedia.org/wiki/Wikipedia:Copyrights}} (CC BY-SA). Therefore, these data can be re-engineered to construct the Complex-TR dataset. Besides, we also engaged college-educated human annotators to verify our test data. We offer the annotators competitive compensation with more than 22 USD in hourly pay, which is significantly higher than the local minimum wage. We also open-source our data and code under the CC BY-SA license. Complex-TR is meant for academic research of LLMs' temporal robustness and reasoning capabilities. However, the retrieved content from Wikipedia and Google Custom Search may contain inappropriate language. Besides, our data augmentation method is based on fictional entities generated by the free version of the ChatGPT\footnote{\url{https://chat.openai.com/}} model. Some of the fictional entities may overlap with real entity names or people's names by coincidence. The generated entities are purely used to improve the temporal reasoning and robustness of LLMs. The authors of this paper hold neutral views toward the generated entities and the contents retrieved from Google Custom Search. 

\bibliography{custom}
\appendix

\begin{table*}[ht]
\small
\resizebox{\textwidth}{!}{
\begin{tabular}{p{0.5\textwidth}p{0.1\textwidth}p{0.3\textwidth}p{0.2\textwidth}}
\toprule
\textbf{ReasonQA Context}                                                                                                      & \textbf{Reason Type} & \textbf{Example Question}                                                                                     & \textbf{Answers}                                    \\
\midrule
Layla Moran held the position of Member of the 57th Parliament of the United Kingdom from June 2017 to November 2019. & L2 1-Hop       & Where was Layla Moran educated in November 2005?                                                     & Brunel University                          \\
\cmidrule(lr){2-4}

Layla Moran studied at UCL Institute of Education from September 2007 to September 2008.                              & L2 M-Hop       & Where was Layla Moran educated from May 2003 to July 2006                                           & Imperial College London, Brunel University \\
\cmidrule(lr){2-4}

Layla Moran held the position of Member of the 58th Parliament of the United Kingdom from December 2019 to May 2023.  & L2 M-Hop       & Where was Layla Moran educated 6 years and 2 months after May 2002?                                  & UCL Institute of Education                 \\
\cmidrule(lr){2-4}

Layla Moran studied at Brunel University from September 2005 to March 2007.                                           & L3 1-Hop       & Where was Layla Moran educated before she studied at Brunel University?                           & Imperial College London                    \\
\cmidrule(lr){2-4}

Layla Moran studied at Imperial College London from September 2000 to August 2003.                                    & L3 M-Hop       & Where was Layla Moran educated 4 years and 11 months after he/she studied at Imperial College London & UCL Institute of Education    \\
\bottomrule
\end{tabular}}
\caption{An example of a ReasonQA context, where the subject is \textbf{Layla Moran}. All information in the \textbf{ReasonQA Context} column is provided to the model along with the question. For the ODQA experiments in our paper, the context will be changed to the Wikipedia article of the subject and the web-retrieved articles based on the question.}
\label{tab:reasonqa-explanation}
\end{table*}

\begin{table*}[t]
\centering
\resizebox{0.8\textwidth}{!}{
\begin{tabular}{lcccccc}
\toprule
                & \multicolumn{2}{c}{\textbf{Single Hop}} & \multicolumn{2}{c}{\textbf{Multi-hop}} & \multicolumn{2}{c}{\textbf{Overall}} \\
                & \textbf{Set Acc.}        & \textbf{Ans. F1}       & \textbf{Set Acc.}       & \textbf{Ans. F1}       & \textbf{Set Acc.} & \textbf{Ans. F1}      \\
\midrule
\textbf{Baseline}         & 46.9           & 47.8          & 29.0          & 37.5          & 36.8         & 42.1         \\
\textbf{Contriever}       & 45.5           & 46.7          & 26.9          & 35.4          & 35.0         & 40.3         \\
\textbf{Contriever-MSMARCO} & 47.6           & 48.7          & 30.7          & \textbf{39.4}          & 38.0         & 43.4         \\
\textbf{GTE}             & \textbf{50.4}           & \textbf{51.1}          & 29.0          & 36.3          & 38.3         & 42.7         \\
\textbf{BGE}             & 49.0           & 49.7          & \textbf{31.2}          & 39.1          & \textbf{38.9}         & \textbf{43.7}   \\
\bottomrule
\end{tabular}}
\caption{Comparison of sentence embedding models for context refinement. The baseline model refers to the \textbf{FiD-PIT} model with the T5-large encoder and simply takes the top 100 passages as context.}
\label{tab:reranker}
\end{table*}

\section{Implementation Details}
\label{sec:implementation}
For the fine-tuning experiments, we used NVIDIA-V100 GPUs. The experiments on \textbf{T5-L} were conducted on 1 40GB-A100 GPU and experiments on \textbf{T5-B} were conducted on 1 12GB-Titan X GPU.  For the \textbf{FLAN-T5-XL} experiments, inference was conducted on 1 40GB-A100 GPU. For the \textbf{T5-B} experiments, the number of training epochs for \textbf{PIT} was 10 and for \textbf{SFT} 15. For \textbf{T5-L} experiments, the number of training epochs was 10 for both \textbf{PIT} and \textbf{SFT}. For the implementation of the FiD model, we followed the GitHub repository of ATLAS\footnote{\url{https://github.com/facebookresearch/atlas}}~\citep{izacard2022few}. This is because this implementation is compatible with later transformer versions and includes many memory-saving functions, such as gradient checkpointing, which suits our computation budget. We use the top 100 ($k=100$) paragraphs for the FiD experiments. The learning rate was set to 1e-5 for all fine-tuning experiments. For the questions with multiple answers, we join all the answers by a special connector ``and''. The predicted string of the models is also split by this connector. For the prompting experiments on \textbf{GPT-3.5} and \textbf{GPT-4}, we used the OpenAI API. The estimated total cost for reproducing all our experiments is 120 USD.  

\section{Problem Settings}
\label{sec:prob-setting-example}

In this section, we elaborate on the two problem settings in detail. In Table~\ref{tab:reasonqa-explanation}, we show an example of a ReasonQA context. The example is about the politician Layla Moran and all related temporal knowledge of Layla Moran is provided as context. In this scenario, a human is able to perform temporal reasoning easily by searching for answers based on the time constraints in the questions. This capability should not be affected by the change of time points. However, in Table~\ref{tab:year-analysis}, we see that LLMs have large performance variations over the years. The results on future years are generally worse than average. However, constantly updating LLMs with new data can be costly and susceptible to catastrophic forgetting. Therefore, it is crucial to adapt LLMs to unseen temporal expressions, e.g., future year tokens. 

On the other hand, the context we used for the ODQA setting is the Wikipedia article on the subject. In a more general open-domain QA setting, models can leverage off-the-shelf retrieval modules to extract more relevant contexts from all over the web. However, the focus of our work is to study the temporal reasoning capability and robustness of LLMs on the temporal QA task. Hence, we leave the open-domain QA setting for future research.

\begin{table}[ht]
\resizebox{\columnwidth}{!}{
\begin{tabular}{lcccc}
\toprule
            & \textbf{L2 1-Hop} & \textbf{L2 M-Hop} & \textbf{L3 1-Hop} & \textbf{L3 M-Hop} \\
            & \textbf{Set Acc.} & \textbf{Set Acc.} & \textbf{Set Acc.} & \textbf{Set Acc.} \\
\midrule
\textbf{T5-B-PIT-SFT}     & 87.0     & 71.9     & 90.5     & 69.1     \\
\textbf{-Resampling} & 85.1     & 69.7     & 88.7     & 66.7     \\
\textbf{-Fictional}  & 83.9     & 68.6     & 87.9     & 65.5   \\
\midrule
\textbf{T5-L-PIT-SFT}     & 91.3 & 72.7 & 91.1 & 71.9 \\
\textbf{-Resampling} & 89.8 & 71.1 & 89.3 & 69.8 \\
\textbf{-Fictional}  & 88.7 & 69.8 & 87.7 & 68.4 \\
\bottomrule
\end{tabular}}
\caption{Ablation studies of our \textbf{T5-B-PIT-SFT} and \textbf{T5-L-PIT-SFT} models on the validation set of Complex-TR in the ReasonQA setting.}
\label{tab:ablation}
\end{table}
\section{Ablation Studies}
\label{sec:ablation}
\subsection{Pseudo-Instruction Tuning}
In Table \ref{tab:ablation}, we show the ablation studies of pseudo-instruction tuning (\textbf{PIT-SFT} models). We examine the two components of \textbf{PIT} in the ReasonQA setting. The first is temporal resampling, and the second is the usage of fictional names. We created two other pseudo-instruction training sets of the same size and examined the final \textbf{PIT-SFT} performance. From Table \ref{tab:ablation}, we can see that removing temporal resampling leads to 2.2 set accuracy drop for L2 M-hop questions and 2.4 for L3 M-hop questions (for the \textbf{T5-B} model). This shows that for the same amount of data, emphasizing the data of low-frequency years leads to better performance. On the other hand, if we use real-world data with shifted temporal information, the performance drop is significant. For the \textbf{T5-B-PIT-SFT} model, changing the fictional names to real-world data can lead to 3.3 and 3.6 set accuracy drop for L2 and L3 multi-hop questions. This performance drop could have resulted from the lack of entity diversity from the temporally-shifted data.

\subsection{Comparison of Context Refinement Models}
Since the context length of our ODQA experiment exceeds the limit of most LLMs, the context refinement process is highly important for the performance of ODQA. In this section, we show the performances of different sentence embedding models for the ODQA setting. We experimented with several popular sentence embedding models for information retrieval and leading open-source models on the Massive Text Embedding Benchmark (MTEB, \citealp{muennighoff-etal-2023-mteb}). 

The models include: (1) \textbf{Contriever}~\citep{izacard2022unsupervised} This model is a popular embedding model trained on contrastive learning in an unsupervised manner.  (2) \textbf{Contriever-MSMARCO}~\citep{izacard2022unsupervised} This model is the contriever model further fine-tuned on the massive MS MARCO~\citep{bajaj2016ms} dataset, which drastically improved the information retrieval performance of contriever. (3) \textbf{GTE} The General Text Embedding model was trained by multi-stage contrastive learning and demonstrated strong performance in various sentence embedding tasks, such as MNLI.  (4) \textbf{BGE} \cite{bge_embedding} This model is short for BAAI General Embedding. It includes a family of sentence embedding models. It demonstrates strong performance on the MTEB leaderboard and we used \texttt{bge-base-en-v1.5} as our context refinement model.

In Table~\ref{tab:reranker}, we show the ODQA experiments of the FiD-PIT-Large model with different embedding models as the re-rankers. Interestingly, we can see that the embedding models affect single-hop and multi-hop performance differently. Contriever-MSMARCO and BGE show a more positive impact on the multi-hop questions whereas GTE has the best performance for single-hop temporal questions. In terms of overall performance, BGE is the best among all the sentence embedding models.

\begin{table}[ht]
\resizebox{\columnwidth}{!}{
\begin{tabular}{lcccc}
\toprule
             & \textbf{Set Acc.} & \textbf{Ans. F1} & \textbf{EM}   & \textbf{Tok. F1} \\
\textbf{T5-B-FiD}$^\dagger$     & -       & -       & 10.3 & 19.7    \\
\midrule
\textbf{T5-B}         & 32.9    & 34.9    & 37.3 & 46.8    \\
\textbf{T5-B-PIT}     & 34.2    & 36.4    & 39.0   & 48.4    \\
\textbf{T5-B-FiD}     & 39.4    & 41.7    & 44.3 & 53.2    \\
\textbf{T5-B-FiD-PIT} & 41.1    & 43.3    & 46.0 & 54.7    \\
\textbf{T5-L-FiD}     & 45.1    & 47.6    & 50.5 & 59.8    \\
\textbf{T5-L-FiD-PIT} & 47.3    & 49.8    & 52.7 & 61.0   \\
\bottomrule
\end{tabular}}
\caption{Experiments on fine-tuning on TimeQA-Hard. We follow the default OBQA setting of their paper. Results with $^\dagger$ are taken from \citet{chen2021dataset}. }
\label{tab:timeqa-exp}
\end{table}

\section{TimeQA Experiments}
\label{sec:further-exp}

In this section, we evaluate the \textbf{PIT} strategy on prior temporal QA datasets. We used \textbf{T5-B-PIT} as initialization, and then fine-tuned on the relevant task datasets for the experiments in this section. In Table~\ref{tab:timeqa-exp}, we show the experimental results of \textbf{PIT} on TimeQA \citep{chen2021dataset}. We only conducted experiments on the Hard subset since the temporal reasoning involved in the Easy subset is too simple. Fusion-in-Decoder (FiD) (\citealp{izacard-grave-2021-leveraging}) was used as the baseline method for TimeQA. Our re-implemented FiD baseline achieved significantly higher results compared to the results of \citet{chen2021dataset}. We can see that \textbf{PIT} is able to improve the performance of both conventional T5-SFT and the more advanced FiD model. This result indicates that the improved temporal reasoning capability of PIT can be transferred to the downstream open-book QA setting, and confirms the generalizability of the PIT method.

\begin{table*}[ht]
\begin{tabular}{p{0.3\textwidth}p{0.1\textwidth}p{0.6\textwidth}}
\toprule
\textbf{Property}                       &  \textbf{Type} & \textbf{Template}                                       \\
\midrule
\textbf{P54} \textit{member of sports team}      &      L2 M-hop          & Which team did \textbf{subject} play for from $t_{1}$ to $t_{2}$?             \\
\midrule
\textbf{P39} \textit{position held}             &       L2 M-hop         & Which position did \textbf{subject} hold $\Delta t$ before $t_{1}$           \\
\midrule
\textbf{P108} \textit{employer}                  &    L3 M-hop            & Which employer did \textbf{subject} work for $\Delta t$ after he/she studied at \textbf{object?}         \\
\midrule
\textbf{P102} \textit{member of political party} &        L2 1-hop        & Which political party did \textbf{subject} belong to in $t_{1}$? \\
\midrule
\textbf{P286} \textit{head coach}                &        L2 M-hop        & Who was the head coach of \textbf{subject} from $t1$ to $t2$?          \\
\midrule
\textbf{P69} \textit{educated at}                &      L3 M-hop      & Where was \textbf{subject} educated when he/she was living in \textbf{object}?              \\
\midrule
\textbf{P488} \textit{chairperson}               &      L2 M-hop          & Who was the chair of \textbf{subject} $\Delta t$ before $t_{1}$?              \\
\midrule
\textbf{P6} \textit{head of government}          &     L2 M-hop           & Who was the head of \textbf{subject} from $t_{1}$ to $t_{2}?$             \\
\midrule
\textbf{P35} \textit{head of state}              &     L3 1-hop           & Who was the head of state of \textbf{subject}  after \textbf{object}'s term of head of state?\\
\midrule
\textbf{P127} \textit{owned by}                  &      L3 M-hop          & Who was the owner of \textbf{subject} when \textbf{object} was the chair?               \\
\midrule
\textbf{P26} \textit{spouse}                     &          L3 M-hop      & Which team did \textbf{subject} play for when he/she was married to \textbf{object}               \\
\midrule
\textbf{P166} \textit{award received}            &          L3 M-hop      & Which award did \textbf{subject} receive when he/she was working for \textbf{object}?             \\
\midrule
\textbf{P937} \textit{work location}             &       L3 M-hop         & Where did \textbf{subject} work when he/she was married to \textbf{object}                     \\
\midrule
\textbf{P551} \textit{residence}                 &      L2 1-Hop          & What was the residence of \textbf{subject} in $t_{1}$\\
\bottomrule
\end{tabular}
\caption{Example templates for Wikidata properties for our Complex-TR dataset.}
\label{tab:question-templates}
\end{table*}

\begin{table*}[ht]
\small
\begin{tabular}{p{0.1\textwidth}p{0.1\textwidth}p{0.8\textwidth}}
\toprule
\textbf{Types}     & \textbf{Number} & \textbf{Examples}                                                                                                                                                 \\
\midrule
countries & 100    & Unatin, Lislands Ofnited, Dencuslandsand, Djisvalwan, New Saintco Moazer, Nuazbe  \\
\midrule

companies & 240    & BrightBoost, AzureAlly, VitalVisionary, LuminaryLogic, NightOwl, AquaAdventures   \\
\midrule

teams     & 152    & Polar Bears, Ice Warriors, Ice Breakers, Polar Storm, Arctic Foxes, Blizzard, Snow Leopards                                     \\
\midrule

towns     & 126    & Fluoriteville, Galenaville, Heliodorhill, Iolitetown, Jadebrook, Kyaniteville, Labradoritehill                                        \\
\midrule

people    & 3,000   & Angelina Romito, Matthew Thompson, Clifford Jump, Barbara Martinez, Martin Dudley, Joseph Parker, Harry Hatch, Richard Driskell, Catherine Scianna       \\
\midrule

schools   & 224    & Yellowwood College, Azura University, Bluebell College, Cactus University, Daybreak College \\
\midrule

awards    & 86     & Masterful Memoirist Medal, Stellar Science Fiction Story Award              \\
\bottomrule
\end{tabular}
\caption{Examples of fictional entities used for pseudo-instruction tuning (PIT). The fictional names are obtained by conversation with the free version of ChatGPT.}
\label{tab:fiction-entities}
\end{table*}

\begin{figure}[ht]
\centering
        \resizebox{1\columnwidth}{!}{
        \begin{tabular}{p{10cm}}
            \toprule
            \textit{\textbf{Example 1}}\quad\quad\\
            \textbf{Error Cause}: Misunderstanding of temporal overlap. 
            
            \textbf{Question}: Which employer did Hans Kramers work for in September 1931?
            \quad\quad

             \textbf{Context}: 
            Hans Kramers worked for: 
            
            \textbf{Leiden University} from January 1934 to January 1952. 
            
            \textbf{Utrecht University} from January 1926 to January 1934.
            
            \textbf{Delft University of Technology} from January 1931 to January 1952.
            
            ......
             \\
             %\textbf{GPT-4's Prediction}: \colorbox{LightGreen}{\textbf{Delft University of Technology}} and \colorbox{LightRed}{\textbf{Leiden University}} 
            \textbf{GPT-4's Prediction}: \textbf{Delft University of Technology} and \textbf{Leiden University}
             \\
             %\textbf{Ground Truth}: \colorbox{LightGreen}{\textbf{Delft University of Technology}} and \colorbox{LightYellow}{\textbf{Utrecht University}} 
             \textbf{Ground Truth}: \textbf{Delft University of Technology} and \textbf{Utrecht University}
             \\
             \bottomrule
        \end{tabular}}
        \resizebox{1\columnwidth}{!}{
        \begin{tabular}{p{10cm}}
            \toprule
            \textit{\textbf{Example 2}}\quad\quad\\
            \textbf{Error Cause}: Misunderstanding of temporal containment. 
            
            \textbf{Question}: Which employer did Elon Musk work for 3 years and 6 months before he/she was living in Boca Chica (Texas)?
            \quad\quad

             \textbf{Context}: 
            Elon Musk worked for: 
            
            \textbf{OpenAI} from December 2015 to January 2019.
            
            \textbf{SpaceX} from June 2002 to Oct 2023. 

            \textbf{Neuralink} from July 2016 to Oct 2023.

            \textbf{The Boring Company} from December 2016 to Oct 2023.
            
            \textbf{Tesla Inc.} from April 2004 to Oct 2023.
            
            ......

            Elon Musk lived in: 
            
            \textbf{Boca Chica (Texas)} from June 2021 to Oct 2023. 
            
             ......
             \\
             \textbf{GPT-4's Prediction}: \textbf{The Boring Company} and \textbf{Neuralink}
             \\
             \textbf{Ground Truth}: \textbf{The Boring Company} and \textbf{Neuralink} and \textbf{OpenAI} and \textbf{Tesla Inc.} and \textbf{SpaceX.}
             \\
             \bottomrule
        \end{tabular}}
    \caption{Examples of GPT-4's erroneous temporal reasoning in the ReasonQA setting.}
    \label{fig:error-analysis-1}
    \end{figure}

\section{Error Analysis}
\label{sec:error-analysis}

In this section, we aim to analyze some mistakes by LLMs in temporal reasoning. We mainly investigate the errors made by GPT-4, because this model has demonstrated excellent performance on various professional and academic benchmarks. We find that GPT-4 still makes mistakes in temporal reasoning. In Figure~\ref{fig:error-analysis-1}, we can see that in Sept. 1931, Hans Kramers was working for both Delft University of Technology and Utrecht University (January 1926 -- January 1934). GPT-4 did a good job of finding the answer Delft University of Technology, but it failed to find the other answer Utrecht University and instead gave the wrong answer Leiden University. This shows that in the multi-answer scenario, GPT-4 can find a good answer but struggles to find all answers. This can also be seen from Table~\ref{tab:compare-metrics}, where GPT-4 has a significantly higher answer-F1 score and a much lower set accuracy score for multi-answer questions.

For the second example in Figure~\ref{fig:error-analysis-1}, we need to first find the starting time of Elon Musk living in Boca Chica (June 2021) and perform time deduction with respect to that time point. The inferred time of interest is December 2017. GPT-4 was only able to determine that Elon Musk worked for the Boring Company and Neurallink, perhaps because these starting times are closer to December 2017. On a longer time horizon, Musk has been working for Tesla and SpaceX since the early 2000s, but the model failed to include these companies in its answers.

\begin{table*}[ht]
\small
\resizebox{\textwidth}{!}{
\begin{tabular}{p{0.5\textwidth}p{0.1\textwidth}p{0.3\textwidth}p{0.2\textwidth}}
\toprule
\textbf{ReasonQA Context}                                                                                                      & \textbf{Type} & \textbf{Example Question}                                                                                     & \textbf{Answers}                                    \\
\midrule
Mary Bartlebaugh studied at Quartz College from May 1872 to May 1874. 
 &   L3 M-hop    &   Which employers did Mary Bartlebaugh work for when he/she was studying at Yam University?                                                  &         Synergy Dynamics                 \\
\cmidrule(lr){2-4}

Mary Bartlebaugh worked for Synergy Dynamics from May 1869 to May 1872. 
                           &  L2 M-hop      &     Which employer did work for Mary Bartlebaugh from Oct 1888 to June 1897                                      &  Solaris Solutions  \\
\cmidrule(lr){2-4}

Mary Bartlebaugh studied at Yam University from May 1867 to May 1871. 
    &    L3 M-hop    &       Where was Mary Bartlebaugh educated when he/she was working for Synergy Dynamics?                          &     Yam University             \\
\cmidrule(lr){2-4}

Mary Bartlebaugh worked for Solaris Solutions from May 1887 to May 1899. 
                                         &    L2 1-hop    &   Where was Mary Bartlebaugh educated at in June 1873?                      &   Quartz College                  \\
\bottomrule
\end{tabular}}
\caption{A pseudo-data example, where the subject is a fictional name, \textbf{Mary Bartlebaugh}.}
\label{tab:example-fictional}
\end{table*}
\begin{figure*}[ht]
    \centering
    \resizebox{\textwidth}{!}{
    \includegraphics{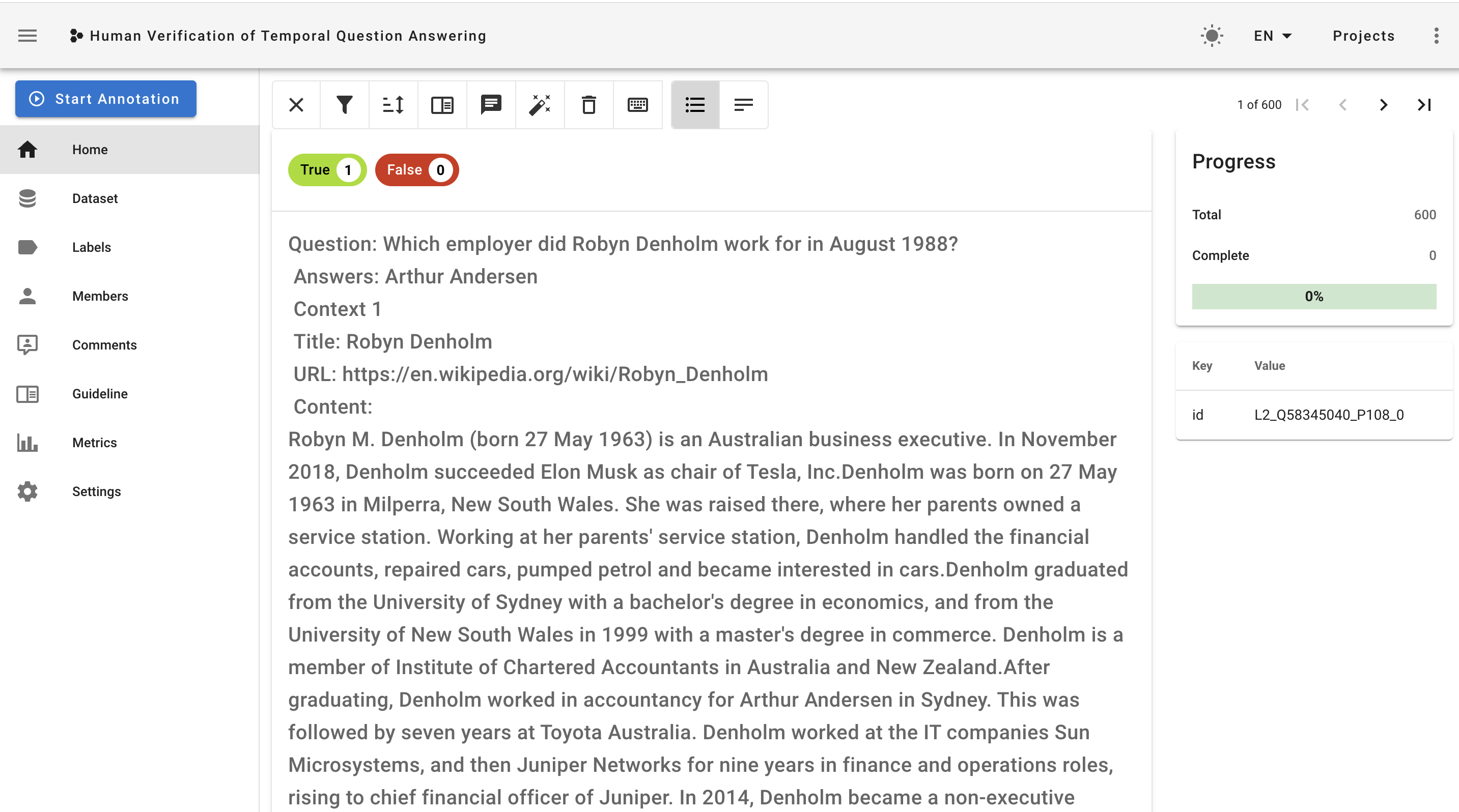}}
    \caption{Annotation interface for the human verification process. Annotators are only asked to give True or False labels to the QA pairs and their contexts.}
    \label{fig:annotation-interface}
\end{figure*}
%\section{Pseudo Instruction for In-context Learning}

\section{Question Templates}
\label{sec:question-templates}

In this section, we show examples of our templates to create the Complex-TR dataset. We used 14 temporally related properties in the Wikidata KB. An example template for each property is shown in Table~\ref{tab:question-templates}.

\section{Examples of Fictional Entities}
\label{sec:fiction-explanations}

In Table~\ref{tab:fiction-entities}, we show some example fictional entities that we used to construct the pseudo-data for \textbf{PIT}. We also show an actual group of fictional data in Table~\ref{tab:example-fictional}.

\section{Annotation Interface}
\label{sec:annotation-interface}

To enhance the accessibility and clarity of the human verification process, we hosted a user-friendly interface on Heroku\footnote{\url{https://www.heroku.com}}. Our interface was built on a popular open-source data annotation Github repository named doccano\footnote{\url{https://github.com/doccano/doccano}}~\citep{doccano}. A screenshot of the annotation interface is given in Figure~\ref{fig:annotation-interface}.
\end{document}